\documentclass{article}
\usepackage{ijcai24}

\usepackage{times}
\usepackage{soul}
\usepackage{url}
\usepackage[hidelinks]{hyperref}
\usepackage[utf8]{inputenc}
\usepackage[small]{caption}
\usepackage{graphicx}
\usepackage{amsmath}
\usepackage{amsthm}
\usepackage{booktabs}
\usepackage{tabularx}
\usepackage{amssymb}
\usepackage{algorithm}
\usepackage{algorithmic}
\usepackage[switch]{lineno}
\usepackage{xcolor}

\linenumbers

\title{Large Language Models for Automatic Milestone Detection in Group Discussions}

\author{
Zhuoxu Duan$^1$
\and
Zhengye Yang$^1$\and
Samuel Westby$^2$\and
Christoph Riedl$^2$\and
Brooke Foucault Welles$^2$\And
Richard J. Radke$^1$
\affiliations
$^1$Rensselaer Polytechnic Institute\\
$^2$Northeastern University
\emails
\{duanz2, yangz15\}@rpi.edu,
\{westby.s, c.riedl, b.welles\}@northeastern.edu,
rjradke@ecse.rpi.edu
}

\begin{document}
\nolinenumbers
\maketitle

\begin{abstract}
    Large language models like GPT have proven widely successful on natural language understanding tasks based on written text documents.  In this paper, we investigate an LLM's performance on recordings of a group oral communication task in which utterances are often truncated or not well-formed.  We propose a new group task experiment involving a puzzle with several milestones that can be achieved in any order. We investigate methods for processing transcripts to detect if, when, and by whom a milestone has been completed. We demonstrate that iteratively prompting GPT with transcription chunks outperforms semantic similarity search methods using text embeddings, and further discuss the quality and randomness of GPT responses under different context window sizes.
\end{abstract}

\section{Introduction}

Group meetings often involve the accomplishment of milestones, i.e., smaller subtasks upon which the group achieves consensus, in order to achieve an overall goal.  However, determining if, when, and by whom milestones have been achieved is currently requires tedious, manual annotation, which is time- and resource-consuming.  The new generation of large language models (LLMs) has the potential to achieve human-level annotation in a matter of seconds, which would be hugely beneficial to communities that study team dynamics, social signal processing, and organizational psychology.  Automated milestone detection could also enable novel applications for human-AI teaming \cite{westby2023collective}, task allocation and scheduling in team-based crowdsourcing \cite{retelny2014expert}, and product development projects \cite{von2006promise}. 

In this paper, we propose a new group task experiment involving a puzzle that has several milestones that can be approached in any order, culminating in a decision that reveals the puzzle's solution.  This task maps well to many process-based theories of group decision making, and allows researchers to investigate interesting questions such as the effect of a facilitator on milestone achievement, the temporal evolution of group performance and individual contributions, and the emergence of group/individual performance and engagement.  

To enable this type of analysis, we investigate methods for automatically processing transcripts of the group meeting to accurately detect if a milestone has been achieved, and the specific participant-tagged utterance at which this occurred.  This is a difficult problem since in our task, groups often propose several incorrect solutions and circle back to different milestones in an unknown sequence. In this way, the task mimics real-world conditions where a solution can be arrived at through many different paths. We begin by considering a baseline algorithm based on sentence embedding before demonstrating the ability of a pre-trained Large Language Model (LLM), namely the recently-introduced ChatGPT framework \cite{openai2023gpt4}, to address the problem. We critically evaluate the performance of such an LLM-based system and show that, while some of the milestones can be detected with perfect accuracy, LLMs can also confidently create false positives that should be taken into consideration, and also have difficulty producing repeatable, well-formatted results.

\section{Related Work} \label{sec:related}

Group performance on tasks has been studied from several angles, covering cognitive \cite{dechurch2010cognitive,hong2004groups}, structural \cite{valentine2015team}, and behavioral \cite{de2012beyond,kim2008meeting} elements.  Particularly germane to our method of automatic milestone detection are scenarios in which consensus must be reached or sub-tasks must be accomplished. In these scenarios, automatic milestone detection could provide researchers and teams with fine-grained analytics, real time progress reports, and a deeper understanding of how consensus and sub-tasks affect overall team effectiveness. 

The automatic analysis of such group tasks proceeds from studying various channels of recorded or online input, which typically include raw audio from each participant, video recordings of the group and the individuals, and/or participant-tagged transcriptions of the meeting.  For example, Lix et al.~\cite{lix2022aligning} recently proposed the concept of discursive diversity to automatically determine the point at which a team reaches consensus, using text from Slack chats. This point is computed based on a moving average of cosine similarities between word2vec embeddings \cite{mikolov2013efficient} of participants' utterances.  Riedl and Woolley \cite{riedl2017burstiness} also use text analysis to detect communication related to different aspects of a task similar to milestones.  Bhattacharya et al.~\cite{bhattacharya-icmi18} proposed several techniques for analyzing multi-modal recordings of the \textit{NASA Moon Survival Task} \cite{hall1970}, including a natural language processing method for detecting topic utterances and rankings. 

Since the performance of these types of algorithms depends on an accurate participant-tagged transcript of a live meeting, natural language processing (NLP) methods play an important role, both in the initial transcription and in the subsequent automatic analysis of the transcript.  Transcription methods are constantly maturing, culminating most recently in transformer-based tools like Whisper from OpenAI, as well as captioning tools built into videoconferencing software like Zoom and Webex.  As we discuss below, while the performance of these tools can be quite good, for audio of unstructured group discussion, transcripts often need to be manually touched up for subsequent use.

Once an accurate transcript has been obtained, general NLP tools for understanding it include information retrieval \cite{manning2009introduction}, semantic similarity search \cite{Chandrasekaran_2021}, and in particular question answering (QA)  \cite{rajpurkar2016squad,hao2017end}, in which natural-language questions are posed with respect to a text corpus, resulting in a natural-language answer or a multiple-choice selection.


Large language models have emerged as a transformative force within the NLP field \cite{brown2020language}. These models, which include LLaMA (Meta), PaLM (Google), BERT (Google) \cite{devlin2018bert}, GPT (OpenAI) \cite{openai2023gpt4}, and their subsequent iterations, are trained on massive textual corpora and have shown remarkable proficiency in understanding and generating natural language. Due to their transformer-based architecture \cite{vaswani2017attention}, they excel at capturing the contextual nuances within text. Their pre-training and fine-tuning methodology enables them to adapt to a broad range of NLP tasks, from translation to text generation, and perhaps even as implicit computational models of humans \cite{horton2023large}.  

Upon the release of GPT4 \cite{openai2023gpt4}, researchers quickly applied it to various problems and soon noticed that its limited memory, i.e., context window, can be an issue. Recursively feeding chunks of text to the model \cite{wu2021recursively} has been a popular method; however, losing information between sessions is likely. Although another ongoing direction aims to extend context windows of models \cite{dong2023survey} to allow longer inputs, this will continue to be an issue in the short term.  \cite{liu2023lost} also observed that long-context models pay less attention to the middle of prompts. To alleviate these issues, \cite{packer2023memgpt} proposed a hierarchical memory system chained with function calls to access memory beyond the context window. All these approaches focus on clean text data from books, Wikipedia, text chatting, and so on, and target consistent conversations with LLMs or evaluate an information retrieval aspect. However, there are substantially fewer studies on LLMs' performance on complex oral communications and their transcripts. 


\section{Methods} \label{sec:Method}
In this section, we will introduce the methods that we investigated to solve the milestone detection problem.

\subsection{The Puzzle Task} \label{sec:puzzle}

\begin{figure}[htbp!]
   \centering
   \includegraphics[width=.8\linewidth]{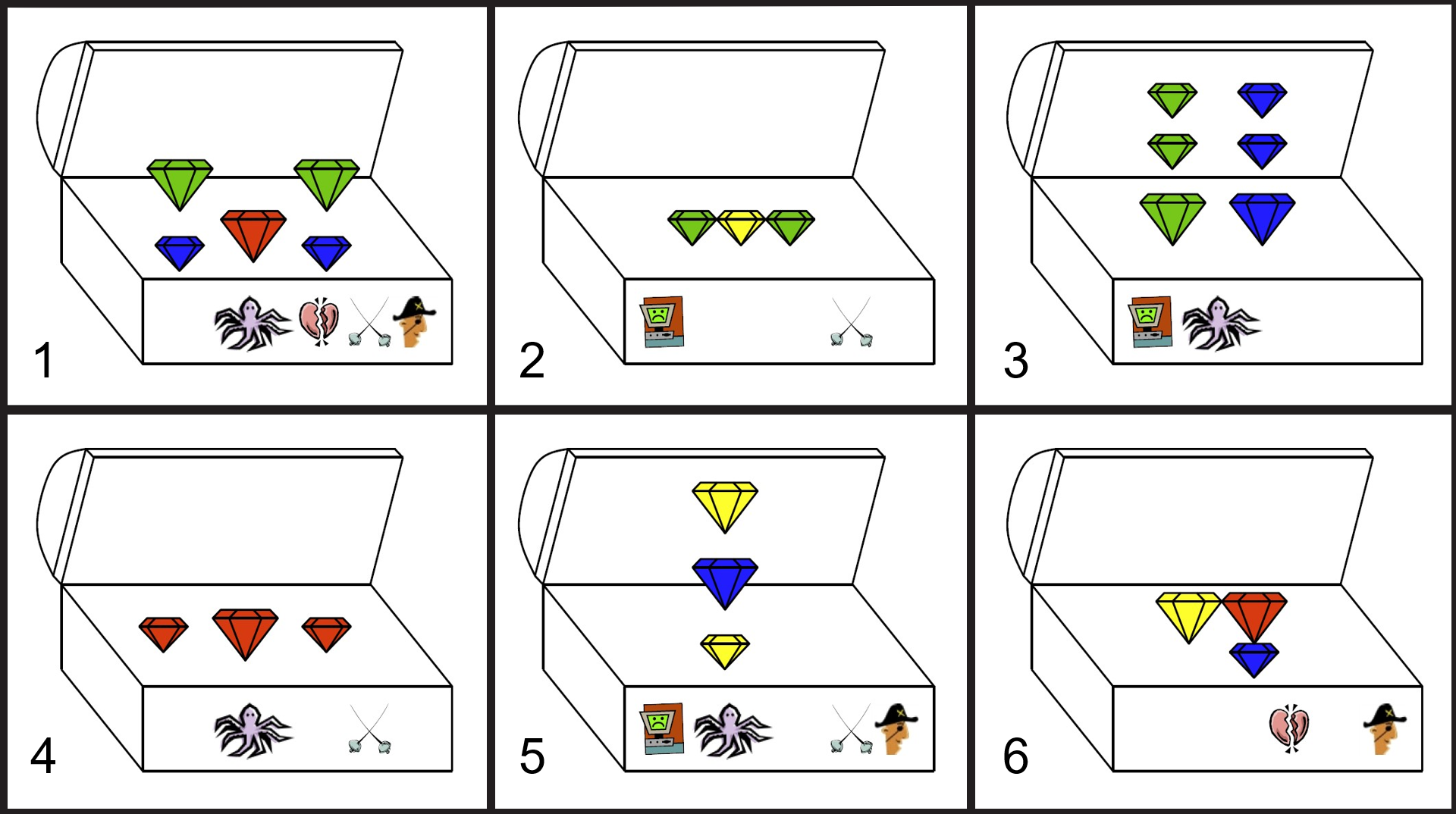} 
   \caption{A subset of the Cursed Treasure puzzle task.}
   \label{image:puzzle}
 \end{figure}

We developed a group discussion task based on a puzzle called ``Cursed Treasure'', designed by Jay Lorch and Michelle Teague\footnote{\url{https://jaylorch.net/static/puzzles/CursedTreasure/CursedTreasure.html}}.  
Participants are given 24 pictures of ``cursed'' treasure chests containing colored gems; a subset of chests is illustrated in Figure \ref{image:puzzle}.

To solve the puzzle, the group must infer the rules that govern when each curse is present.  For example, the ``octopus'' curse is present if and only if no gems are touching, and the ``hex'' curse is present if and only if there are no red gems.   If all five curses are successfully decoded, the group applies the learned rules to a final set of chests to determine a secret solution phrase of the puzzle.  Since the groups can discuss and discover the curses in any order, and there are no explicit instructions, this group task is much more challenging and open-ended than classical group-study instruments like the Moon Survival Task, which typically involve  clearly defined phases with easily identifiable milestones. While it is not germane to this paper, the experiment featured a recorded ``Puzzle Master'' that gives the groups timed hints that may or may not be useful.  

\subsection{Dataset Acquisition and Preprocessing}

We invited 20 groups of three to four people to attempt the puzzle task via Zoom meetings under an approved IRB protocol.  Each participant used a laptop configured by the experimenters to access the meeting room, and each group had 40 minutes to get as far as they could in the task. Video recordings and transcripts of the audio data were obtained from the Zoom application.
Table \ref{tab: achievement} shows the milestones each team solved. Among them, ``one'' and ``hex'' seem challenging, while ``octopus'' and ``quadruple'' are easier. Three teams solved the entire puzzle.

\begin{table*}[htbp!]
  \caption{Milestones achieved by the 20 teams.}
  \label{tab: achievement}
  \small
  \begin{tabularx}{\textwidth}{c*{20}{>{\centering\arraybackslash}X}}
    \toprule
    Team & 1 & 2 & 3 & 4 & 5 & 6 & 7 & 8 & 9 & 10 & 11 & 12 & 13 & 14 & 15 & 16 & 17 & 18 & 19 & 20\\ 
    \midrule
    one         &  & \checkmark & \checkmark &  & \checkmark & \checkmark &  &  &  & \checkmark & \checkmark & \checkmark & \checkmark &  & \checkmark & \checkmark &  &  &  & \checkmark\\
    dual        &  & \checkmark & \checkmark & \checkmark & \checkmark & \checkmark & \checkmark & \checkmark & \checkmark & \checkmark & \checkmark & \checkmark & \checkmark & \checkmark & \checkmark & \checkmark & \checkmark &  & \checkmark & \checkmark\\
    quadruple   & \checkmark & \checkmark & \checkmark & \checkmark & \checkmark & \checkmark & \checkmark &  & \checkmark & \checkmark & \checkmark & \checkmark & \checkmark & \checkmark & \checkmark & \checkmark & \checkmark &  & \checkmark & \checkmark\\
    octopus     & \checkmark & \checkmark & \checkmark & \checkmark & \checkmark & \checkmark & \checkmark & \checkmark & \checkmark & \checkmark & \checkmark & \checkmark & \checkmark & \checkmark & \checkmark & \checkmark & \checkmark & \checkmark & \checkmark & \checkmark\\
    hex         &  & \checkmark & \checkmark &  & \checkmark & \checkmark & \checkmark &  & \checkmark & \checkmark & \checkmark & \checkmark & \checkmark &  & \checkmark &  &  & \checkmark & \checkmark & \checkmark\\
    solution    &  &  &  &  &  & & & &  & \checkmark & \checkmark & \checkmark & &  &  &  &  &  &  & \\
    \bottomrule
\end{tabularx}
\end{table*}

Due to the intensive discussion during the meetings, cross-talk is common in the audio data. This led to speech-to-text transcription issues since utterance detection and speaker recognition are difficult. Consequently, the automatic Zoom transcriptions tended to be short sentence fragments with partially incorrect speakers, and most cross-talk sections were unrecognized. Furthermore, keywords are frequently wrong in the transcription, such as mistaking ``jam'' for ``gem'' and ``obstacle'' for ``octopus''.

\begin{figure}[!]
  \centering
  \includegraphics[width=\linewidth]{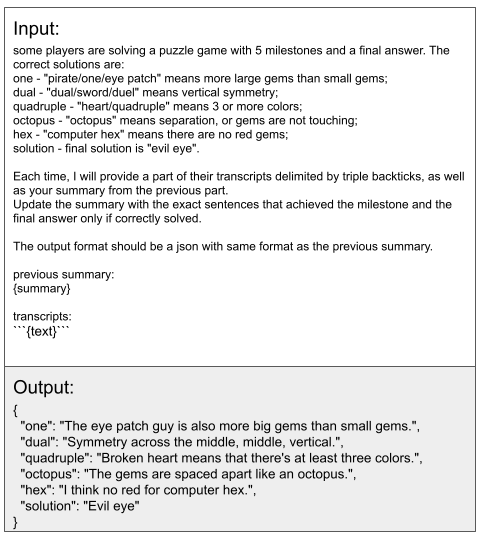} 
  \caption{An example of our milestone detection prompt to ChatGPT (not including the full transcription of the group meeting in this figure).  In the structured output, ``one, dual, quadruple, octopus, hex'' are the five milestones for the individual curses, and ``solution'' is the final answer. In this example, ChatGPT accurately found all of the correct milestone sentences.}
  \label{image: promptDemo}
\end{figure}

To alleviate such issues, the audio data was also transcribed using the speech-to-text service Rev.com, which provides transcriptions using AI with human correction. Although this high-quality transcription solved most of the issues, its results tended to group multiple short sentences together into long multi-sentence monologues without precise time stamps.  This introduced another problem since we want to precisely determine the sentence or sentence fragment at which a milestone is achieved.   Therefore, we mapped the Rev content to the shorter Zoom sentence structure using time-alignment-based coding and manual correction.  We used these high-accuracy, short-fragment transcriptions in our subsequent experiments. The Puzzle Master's comments, if present, are not included in the transcriptions.

As ground truth, we manually annotated correct, complete statements of the milestones in these transcriptions.  Discussions about a particular target may contain initial proposals, subsequent discussions or confirmations, and restatements when forming a final solution.  For our purposes, we annotated any sentence that correctly stated a milestone as valid (be it the first time the milestone is achieved or a subsequent restatement/confirmation of the milestone).  While not explicitly included in the tables below, each sentence is explicitly tagged with the name of the person who uttered it, and associated with a wall-clock time stamp.

\subsection{Milestone detection by prompting GPT}

With the public release of ChatGPT (back-ended by GPT-4) in early 2023, researchers quickly realized the potential of this new model for sophisticated natural language understanding and generation \cite{horton2023large}.  Unlike the more traditional embedding-based approaches we will discuss in Section \ref{section4: Experiments}, GPT is interacted with via a natural-language prompt, which turns the problem into prompt engineering to some extent. The goal is to obtain consistent, well-formatted, and correct responses from GPT.

Ultimately, the prompt structure we arrived at is illustrated in Figure \ref{image: promptDemo}.  
The prompt includes a description of the task, the solution to each milestone, the transcript from the group discussion (see more below), and the desired structure for the output, which allows us to automatically parse the results.

However, a key consideration for large language models is the length of their context window and the rate limit for prompting, especially when dealing with extensive text content. The context window can be regarded as the maximum memory length for single prompts and conversation history, which comes from the nature of the model design. The rate limit ensures that individual users do not send too many requests within a period.
At the time of this paper's writing, the production level GPT-4 (gpt-4-0613, the June 13th, 2023 checkpoint) offers 8k and 32k capabilities. Our account only reached tier 1 usage, which allows 10k tokens per minute (TPM). We note that OpenAI recently released GPT-4 Turbo, which has a 128k context window and even larger rates, but it is still in the test phase, and the performance on our task is poor. All the models and limits are listed in Table \ref{tab:gpt_sizes}.

\begin{figure*}[!]
  \centering
  \includegraphics[width=0.8\textwidth]{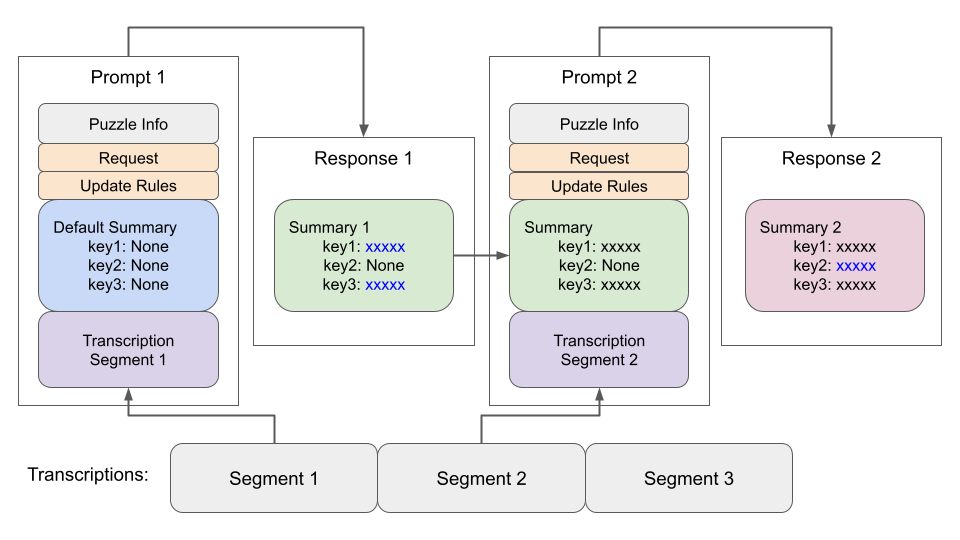}
  \caption{Flowchart of the proposed iterative query. The puzzle information, request, and update rule are fixed in each iteration.  We use a new section of the transcriptions in each loop until all transcriptions are processed. Blue characters in the response window mean GPT has found a better match for that milestone.}
  \label{image: flowchart}
\end{figure*}

\begin{table}
    \centering
    \resizebox{\columnwidth}{!}{
    \begin{tabular}{lrr}
        \toprule
        Model               & Context window(tokens)   & Rate limit(TPM) \\
        \midrule
        gpt-4               & 8,192                     & 10,000        \\
        gpt-4-32k           & 32,768                    & 10,000        \\
        gpt-4 turbo         & 128,000                   & 150,000        \\
        \bottomrule
    \end{tabular}}
    \caption{Released GPT4 models and their corresponding context window sizes, with Tier 1 rate limits (up to the end of 2023).}
    \label{tab:gpt_sizes}
\end{table}

Here, the tokenization of input text is achieved by Byte Pair Encoding (BPE) \cite{sennrich2015neural}.  The number of tokens is not the same as the number of words, since each word could be converted to multiple tokens depending on its subwords and rarity. For our dataset, the transcriptions for each $\sim$ 45-minute team meeting require between 4000 and 12000 tokens.
Given that our experiments are not discussion-intensive, this number would be much larger for real-world meetings. Considering the prior knowledge, requirements, and space for handling response errors, we cannot provide the entire transcript at once in practice as illustrated in Figure \ref{image: promptDemo}.

To address this limitation, we 
choose to use a basic iterative prompting scheme (Figure \ref{image: flowchart}).
The idea is widely adopted in the community and similar to \cite{wu2021recursively}. Since our transcripts are linearly distributed plain texts, applying Mem-GPT \cite{packer2023memgpt} would be somewhat equivalent.  After introducing the task and solutions, the prompt provides a request and a summary of previously-detected milestones with rules for how to update the detections.  The summary is similar to a Python dictionary, with the names of the milestones as the keys and empty strings \texttt{""}
as the initial values.  The values for each key will only be updated when GPT finds a better match in the provided chunk of transcription. The response from GPT is required to be a \texttt{json} in the same format. Detailed configurations and performance will be discussed in Section \ref{section4: Experiments}.


\section{Experiments} \label{section4: Experiments}

\subsection{Baseline Approach}
Before illustrating the experiments with GPT, we first introduce the baseline method we use to make a comparison. Finding potential milestones may be viewed as a semantic similarity search task: locating the most relevant sentence from a transcript compared to a given target sentence. The key point is to find a way to encode the transcripts so that context information in the document is mostly preserved.


We choose to adopt the widely-used Bidirectional Encoder Representation from Transformer (BERT) \cite{devlin2018bert} model for sentence embedding \cite{reimers2019sentence}.  Unlike 
word embeddings,
BERT effectively looks at an entire sentence at once, considering the context of different words at different positions. Moreover, with its internal multi-head self-attention mechanism, the transformer block introduces more capacity to better distinguish differences between dissimilar words. 

One challenge specific to our task, but common in many organizational settings \cite{tasselli2020bridging}, is that groups typically develop a shorthand vocabulary for the milestones that may not match our ground truth solution sentences.  For example, one group may refer to the ``pirate'' while another may refer to the ``eye patch''.   We attempted to address this issue by creating several different synonyms and paraphrases for each milestone (e.g., ``The eye patch means more big diamonds than small ones''; ``Pirate is more large than small'') and computing the similarity score for each solution/candidate pair, selecting the highest similarity as the best match.  We then rank all the candidate sentences for each milestone in order from highest to lowest similarity, and threshold this list to get a best-to-worst set of possible matches, as discussed more below.

\subsection{Evaluation Method}

\begin{figure}[htbp!]
  \centering
  \includegraphics[width=\linewidth]{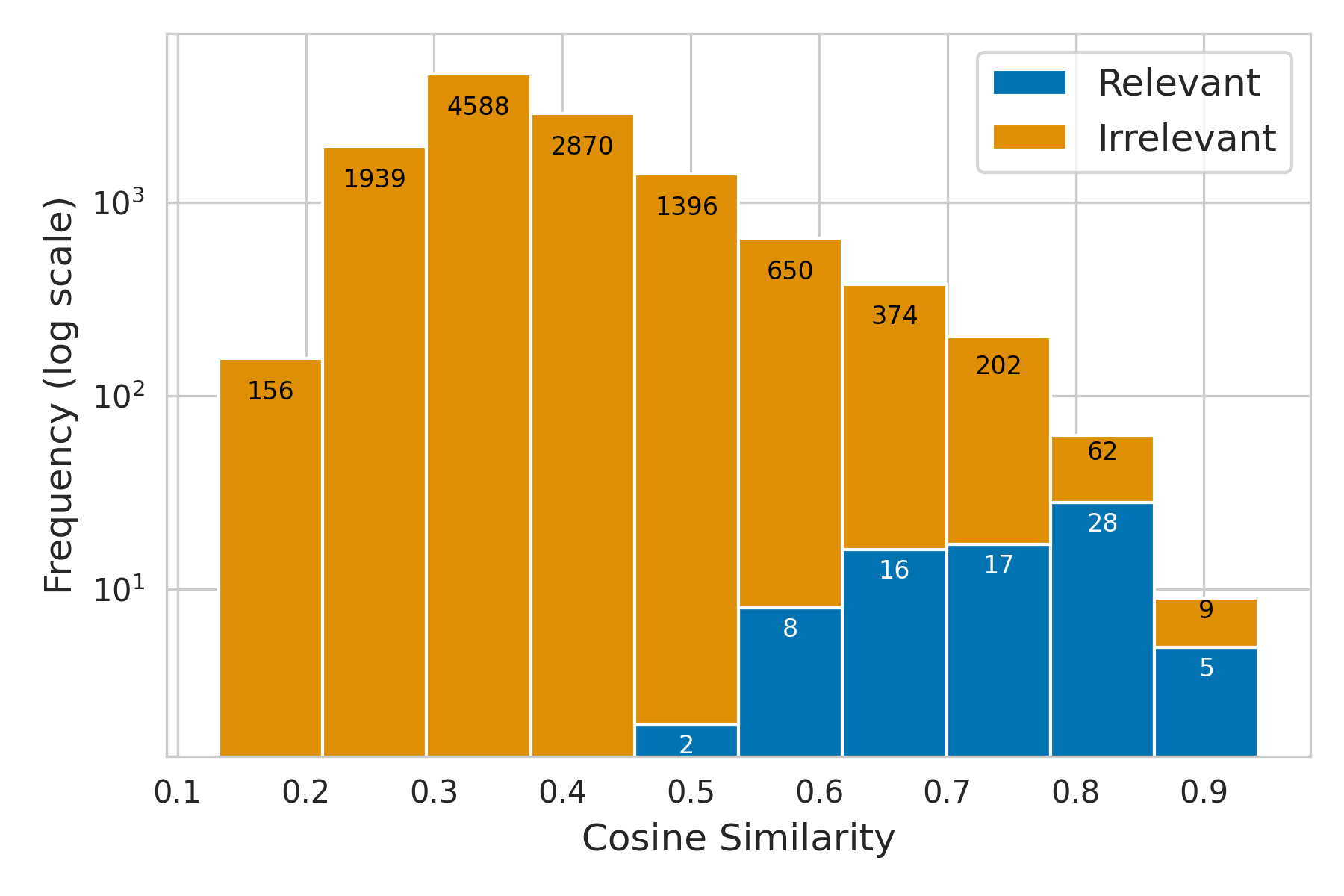}
  \caption{Frequency histogram of cosine similarity for ``octopus". The x-axis is the score ranging from 0 to 1, and the y-axis is the log-scaled frequency. Relevant (milestone) and irrelevant sentences (background) are represented in blue and yellow, respectively.}
  \label{image: histogram}
\end{figure}

For the baseline method, we adopt the \emph{top-k performance} as our evaluation metric.  That is, the method produces a list of matches between candidate sentences and the solution for a given milestone, sorted by decreasing similarity in the embedding space.  We adopted cosine similarity as the sentence-level metric for BERT. If the correct candidate sentence corresponding to a milestone is in the first $k$ entries of this sorted list, we consider it to be a detection (i.e., true positive) at rank $k$.  Thus, rank-1 performance measures how often the correct candidate sentence is at the top of the sorted list.  It is worth noting that the embedding-based method is somewhat ill-posed since our dataset is extremely imbalanced. As Figure  \ref{image: histogram} shows, irrelevant sentences can also have high similarities, and the number is larger than the relevant ones.

\begin{figure}[htbp!]
  \centering
  \includegraphics[width=\linewidth]{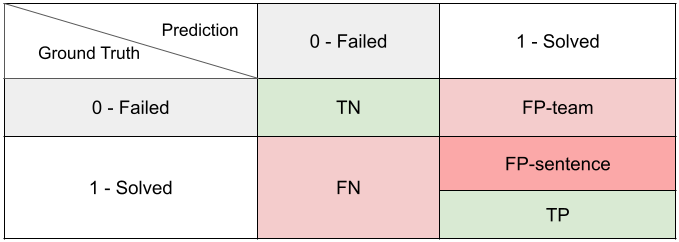}
  \caption{Confusion matrix of the proposed evaluation method. Each team is labeled as failed (0) or solved (1) for each milestone, which produces the team-level true negatives, false negatives, and false positives. When the prediction and ground truth say a team has solved a milestone, the proposed sentences must be checked to determine whether they are the true answers. If not, the situation is considered as a sentence-level false positive.}
  \label{image: confusionMatrix}
\end{figure}

Furthermore, if a team doesn't solve a particular milestone, all of the top $k$ proposals will be wrong.  Thus, we must introduce a threshold for each milestone, to determine whether the match quality is sufficient to declare that the milestone may be present at all.  Based on this framework, Figure \ref{image: confusionMatrix} illustrates the situations that may occur.  If the team did not actually achieve the milestone and no candidate sentences passed the threshold, we consider this a true negative (TN).  If the team achieved the milestone but no candidate sentences passed the threshold, we consider this a false negative or miss (FN).  

Here we note there are two types of false positives that can occur.  The team may fail to achieve the milestone at all, but the algorithm may generate candidate sentences that are above the threshold.  We define this situation as a false positive at the team level (FP-team).  On the other hand, the team may achieve the milestone, but none of the algorithm's proposals above the threshold at rank $k$ may be correct (i.e., the algorithm asserts that the milestone has been found, but the corresponding sentence is incorrect).  We define this situation as a false positive at the sentence level (FP-sentence).  The desired scenario is that a team achieves the milestone, and the algorithm correctly identifies the exact sentence at which the milestone was achieved in its rank $k$ list, which we denote as a true positive (TP).

The final accuracy of a given algorithm involves both team and sentence-level false positives, as shown below:

\begin{displaymath}
    accuracy = \frac{TN + TP}{FP_{t} + FP_{s} + FN + TN + TP}
\end{displaymath}

We note that the top-$k$ evaluation only really applies to the baseline BERT method, since as described in the previous section, our GPT prompting approach expects only the top-1 candidate sentence per milestone. 
Due to the limited size of our dataset, we cannot train the BERT model. Therefore, we apply a pre-trained sentence BERT model and only do zero-shot prompting for GPT to make a fair comparison.





\subsection{Baseline Results}

Using the manually chosen thresholds, BERT's performance is shown in Table \ref{table:BERT}. In addition to accuracy, we also show the counting mistakes made among the 20 teams (FP and FN).
Most of the mistakes come from false positive sentences, which are an inevitable problem in our dataset. As shown in Figure \ref{image: histogram}, 
exception cases will easily pollute the proposals even though the proposed methodology can prioritize relevant sentences. For example, short sentences make it easy to get high scores. The ``dual'' and ``hex'' curses suffer from such situations because their candidate query sentences have short examples: ``The swords mean symmetry''; ''Hex is no red''. However, ``one'', ``quadruple'', and ``octopus'' are more affected by the contextual synonyms.
False positives at the sentence level become fairly low, especially at higher ranks.  We also observe that the false negative counts are close or equal to zero because we chose a low threshold, resulting in a larger chance of getting a false positive at the team level. 

\begin{table}
    \centering
   \resizebox{\columnwidth}{!}{
    \begin{tabular}{lrrrrrr}
        \toprule
        & \multicolumn{2}{c}{Top-1} & \multicolumn{2}{c}{Top-5} \\
        \cmidrule(lr){2-3}\cmidrule(lr){4-5}
        Milestone & acc (\%) & FP-s & acc (\%) & FP-s & FP-t & FN \\ 
        \midrule
        one         & 70 & 2  & 80 & 0 & 4 & 0\\
        dual        & 25 & 13 & 70 & 4 & 2 & 0\\
        quadruple   & 65 & 5  & 75 & 3 & 2 & 0\\
        octopus     & 50 & 10 & 85 & 3 & 0 & 0\\
        hex         & 50 & 5  & 60 & 3 & 5 & 0\\
        solution    & 90 & 0  & 90 & 0 & 2 & 0\\
        \bottomrule
        \end{tabular}}
    \caption{BERT performance}
    \label{table:BERT}
\end{table}


\subsection{GPT experiments}

Fortunately, document-level semantic information is well handled in large language models like GPT \cite{vaswani2017attention,brown2020language}. Contextual synonyms can also be understood as long as they are provided in the prompt. 
These advantages enable LLMs to solve what was previously considered an ill-posed problem. After the release of GPT4, we did a test on the browser ChatGPT interface (with gpt-4-0314, 4096 context window at that time), whose results already outperformed the baseline, as shown in Table \ref{table:ChatGPT}.

\begin{table}[htbp!]
    \centering
    \begin{tabular}{lrrrr}
        \toprule
        Milestone & Accuracy (\%) & FP-s & FP-t & FN \\ 
        \midrule
        one         & 70  & 1 & 5 & 0 \\
        dual        & 75  & 1 & 0 & 4 \\
        quadruple   & 100 & 0 & 0 & 0 \\  
        octopus     & 90  & 2 & 0 & 0 \\
        hex         & 85  & 1 & 2 & 0 \\
        solution    & 100 & 0 & 0 & 0 \\
        \bottomrule
        \end{tabular}
    \caption{ChatGPT (gpt-4-0314) single trial performance}
    \label{table:ChatGPT}
\end{table}

Since OpenAI is constantly updating and releasing new models, we next tested version gpt-4-0613 after getting access to the OpenAI API.  All of the following experiments are done in Python with the OpenAI API, with the chat completion temperature set to 0. We started by assuming a 4096-token context window, inherited from the initial tests with the ChatGPT browser interface. Then we moved to the current 8192 window size and further tested with theoretically large enough windows (32k and 128k).

As discussed in Section \ref{sec:Method}, the iterative prompting method requires space for prior knowledge, requirements, and error responses. In our settings, this non-transcription part of the prompt takes roughly 300 to 400 tokens, although severe violations may happen in the responses to make this part exceed 500 tokens.
Therefore, we restricted the number of tokens to 3600 for each chunk when preparing prompts. This logic is the same for all context window sizes.
For each prompt, we update the summary and transcription parts. 
The top three sections of the prompt remain unchanged. GPT will find answers for each key according to the request and update rules, i.e., ``update the milestone only if you find a better match to the true answers''. After receiving the response, the summary in the following prompt will be changed based on it.

Due to the non-deterministic nature of GPT \cite{openai2023gpt4}, we repeated the same prompting for each of the 20 teams over 10 trials to observe the response. Table \ref{table:gpt-4 performance} shows the average performance with standard deviations where the counts of mistakes (FP-s, FP-t, FN) are averages over the 10 rounds.
        
\begin{table}[htbp!]
    \centering
    \begin{tabular}{lrrrrr}
        \toprule
        Milestone   & Accuracy (\%)     & FP-s  & FP-t  & FN    \\
        \midrule
        one         & 88.0 $\pm$ 4.2    & 0.5     & 1.9    & 0.0     \\
        dual        & 90.0 $\pm$ 7.1    & 1.6    & 0.0     & 0.4     \\
        quadruple   & 93.5 $\pm$ 4.7    & 0.3     & 0.0     & 1.0    \\
        octopus     & 94.5 $\pm$ 3.7    & 1.1    & 0.0     & 0.0     \\
        hex         & 88.0 $\pm$ 4.8    & 0.3     & 2.1    & 0.0     \\
        solution    & 97.5 $\pm$ 2.6    & 0.1     & 0.4     & 0.0     \\
        \bottomrule
    \end{tabular}
    \caption{GPT API (gpt-4-0613, pseudo 4096 window) 10-trial average performance}
    \label{table:gpt-4 performance}
\end{table}

We additionally performed five trials for each team with the 8192 token limit. Our expectation was that longer context windows would provide better results since there is no information loss due to summarization.  The results shown in Table \ref{table:gpt-4 8192 performance} are surprising, showing that the longer context window performs worse than the shorter window. A clear pattern is that the larger window brings a more stable response format and fewer accuracy fluctuations. However, the performance of ``one'' decreased significantly (even worse than the BERT baseline) due to many false alarms on the team level. Although only 11 teams solved ``one'', most teams have discussed this milestone. There is a general drop of 2.5\%$\sim$5\% for other milestones.

\begin{table}[htbp!]
    \centering
    \begin{tabular}{lrrrrr}
        \toprule
        Milestone   & Accuracy (\%)     & FP-s  & FP-t  & FN    \\
        \midrule
        one         & 66.0 $\pm$ 4.2    & 1.0   & 5.2   & 0.6     \\
        dual        & 85.0 $\pm$ 6.1    & 3.0   & 0.0   & 0.0     \\
        quadruple   & 90.0 $\pm$ 6.1    & 0.8   & 0.0   & 1.2    \\
        octopus     & 92.0 $\pm$ 2.7    & 1.6   & 0.0   & 0.0     \\
        hex         & 82.0 $\pm$ 2.7    & 0.8   & 2.8   & 0.0     \\
        solution    & 100. $\pm$ 0.0    & 0.0   & 0.0   & 0.0     \\
        \bottomrule
    \end{tabular}
    \caption{GPT API (gpt-4-0613, 8192 window) 5-round average performance}
    \label{table:gpt-4 8192 performance}
\end{table}

\section{Discussion} \label{sec:Discussion}
\subsection{Performance} \label{sec:5.1}

Compared to the BERT baseline, GPT has better results but makes mistakes in different aspects. Based on the ten-trial responses, the most frequent mistake is being distracted by declarative statements with an affirmative tone, even if they are wrong. This accounts for the majority of team-level false positives and some sentence-level ones, especially when the true milestone tends to be a sentence fragment with an unsure tone. In addition, GPT is sometimes lost when dealing with long discussions, resulting in responses that are an unrelated sentence near the correct conclusion. Finally, randomness is a key issue. There are cases when milestone answers are misplaced: e.g., ``octopus'' with the sentence referring to ``hex''. There is also a chance that even though the team has solved a milestone with a clear statement, GPT will return a blank response. These issues happen randomly with no clear patterns.

\subsection{Repeatability, Formatting, and Hallucinations}

We found the responses of GPT to be significantly non-deterministic even after setting the temperature to 0. While the impact of randomness on performance/accuracy is limited, it does introduce variations in the response. None of the 20 teams had identical responses for the 10 rounds. Three teams were consistent in being correct or making the same mistakes. Five teams had significant variations over the ten trials. Their token lengths vary from 5k to 12k, with no clear trends.

In addition to the random mistakes discussed in Section \ref{sec:5.1}, there are issues including violating the required response format and hallucinations.  
According to the prompt we introduced in Section \ref{sec:Method}, the response must be in JSON format, with each milestone showing an exact sentence that solved it. However, gpt-4-0613 interprets this as returning plain text with the JSON punctuations ``\{\} , :''. This might be solved with proper prompts, but other issues are real limitations. Although the response roughly follows the format, GPT sometimes adds additional descriptions to the answer. For example, ``previous summary: \{ xxx \}, updated summary: \{ xxx \}'', or ``The players didn't achieve any new milestones or the final answer in this part of the transcript. Therefore, the summary remains the same: \{ xxx \}''.  Sometimes, instead of returning a single sentence for each milestone, we observed that GPT with the 4096 context window can return a series of discussions.  These issues happen randomly across different teams with no clear pattern, but stably show up for certain teams during different rounds of tests.  This issues make it difficult to reliably, automatically parse GPT's answers.  

Hallucinations have been an issue for GPT since its introduction. Although OpenAI is constantly working to minimize their frequency, some level of hallucination seems inevitable. In our experiments, GPT hallucinates by giving responses with the true answers that we provided as prior knowledge in the prompts. Among the ten-round 200 individual tests we conducted, hallucination happened 10 times, in four teams with frequencies varying from 1 to 4 and their numbers of tokens varying from 5k to 9k. Again, we found no clear pattern.

Although the numerical performance of GPT is satisfying, the issues we found could be fatal in automatic pipelines and real-world applications in which people unquestioningly rely on the responses.  The mistakes GPT currently makes are unpredictable, hard to discover, and hard to fix. 

The newly released GPT4 Turbo provides more control over reducing randomness and following a specific response format (JSON, for example). We made two rounds of tests with the turbo version and observed that the response fully followed the format instructions. However, since it is still a ``preview'' version, the quality of responses is poor, and hallucination is severe. 

\section{Conclusions}

We demonstrated that the current generation of large language models such as GPT can generally be used ``out of the box'' to solve a milestone detection problem that would have been approached only a few years ago with word or sentence embedding methods, and before that only with manual annotation.  We believe that our puzzle task, with its internal milestones, presents an interesting avenue for future study.  Accurate milestone detection could be used to estimate the leaders and key contributors of a group meeting and study the patterns of communication that lead to efficient and effective decision-making.  

We also demonstrate that carefully crafting the prompts used to query the LLM is a key part of the research process. This includes crafting the prompts defining the context in which answers are sought, defining the actual request, and determining how results can be summarized and updated to overcome token length limitations. We show that with carefully crafted prompts, long pieces of text can be broken into multiple segments which can be fed into LLMs in a piece-wise fashion. 

Future versions of ChatGPT and its successors will almost certainly raise the token limit per session, which would remove the need to split the transcript across multiple prompts.  However, our experiments showed this did not always result in performance improvement.  Furthermore, it is possible that for long, dense meetings, transcripts may still be above a given token limit.  Our experiments also revealed that ChatGPT output can be confidently wrong, hard to automatically parse, and inconsistent over multiple trials.

While our study processed meeting transcripts retrospectively, the same approach could work in near real-time provided that online speech transcription was sufficiently accurate (e.g., using tools built into Zoom and similar software, as well as external tools like OpenAI's Whisper \cite{radford2023robust}).  This could enable intelligent agents to facilitate real-time meetings.  We also have yet to explore the richness of the full multimodal meeting recording (i.e., raw audio and video) to analyze tone of voice, speech patterns, gaze directions, facial expressions, body pose, and gestures.


\appendix

\section*{Ethical Statement}

There are no ethical issues related to the analysis of our meeting transcripts, which were acquired under an approved IRB protocol that described how they would be used.  In future real-world applications of LLMs to meeting analysis (especially in real time), all participants would need to be aware that their utterances might be transmitted to, processed by, and used for training a third-party LLM, which might have implications for the type of confidential or sensitive material that could be discussed.

\bibliographystyle{named}
\bibliography{ijcai24}

\end{document}